\pdfoutput=1

\documentclass[11pt]{article}

\usepackage[final]{acl}
\usepackage{graphicx}
\usepackage{amsfonts}	
\usepackage{times}
\usepackage{latexsym}
\usepackage{amsmath}
\usepackage{caption}
\usepackage{tcolorbox}

\usepackage{subfig}
\usepackage{subcaption}

\usepackage[T1]{fontenc}
\usepackage{caption}
\usepackage[utf8]{inputenc}

\usepackage{float}
\usepackage{microtype}
\usepackage{geometry}
\usepackage{array}
\usepackage{booktabs}
\usepackage{multirow}
\usepackage{listings}
\usepackage{tabularray}
\usepackage{tabularx}
\usepackage{tikz}
\usetikzlibrary{shapes.geometric, arrows.meta}

\usepackage{textcomp}

\usepackage{amsmath}
\usepackage{bussproofs}
\usepackage{tcolorbox}
\usepackage{caption}
\usepackage{subcaption}
\usepackage{booktabs}
\usepackage{graphicx}
\usepackage{cuted}
\usepackage{multicol}
\usepackage{anyfontsize}
\usepackage{rotating}
\usepackage{array,makecell,multirow}
\usepackage{wrapfig}
\usepackage{setspace}
\usepackage{enumitem}
\setlist{leftmargin=1mm}
\usepackage{pifont}
\usepackage{booktabs}
\usepackage{subcaption}
\usepackage[capitalise]{cleveref}

\title {InSaAF: Incorporating Safety through Accuracy and Fairness  \\ Are LLMs ready for the Indian Legal Domain?}

\author{
\fontsize{11pt}{11pt}\selectfont
 \makecell{
 $^\clubsuit$Yogesh Tripathi$^1$ \; 
 $^\clubsuit$Raghav Donakanti$^2$ \;  
 $^\clubsuit$Sahil Girhepuje$^1$ \; 
 Ishan Kavathekar$^2$ \\ 
 Bhaskara Hanuma Vedula$^2$ \; 
 Gokul S Krishnan$^1$ \; 
 Shreya Goyal$^3$ \; 
 Anmol Goel$^2$ \\ 
 Balaraman Ravindran$^{1,4}$
 Ponnurangam Kumaraguru$^2$ \
 }
 \vspace{2pt}
\\
\fontsize{7.3pt}{7.3pt}\selectfont
\makecell{
$^1$ Centre for Responsible AI, Indian Institute of Technology Madras, India \\
$^2$ International Institute of Information Technology, Hyderabad, India \\
$^3$ AmexAI Labs, American Express, Bengaluru \\
$^4$ Wadhwani School of Data Science and AI, Indian Institute of Technology Madras, India \\
$^\clubsuit$ Co-first authors 
}
}

\begin{document}
\maketitle

\begin{abstract}

Recent advancements in language technology and Artificial Intelligence have resulted in numerous Language Models being proposed to perform various tasks in the legal domain ranging from predicting judgments to generating summaries. Despite their immense potential, these models have been proven to learn and exhibit societal biases and make unfair predictions. In this study, we explore the ability of Large Language Models (LLMs) to perform legal tasks in the Indian landscape when social factors are involved. We present a novel metric, $\beta$-weighted \textit{Legal Safety Score ($LSS_{\beta}$)}, which encapsulates both the fairness and accuracy aspects of the LLM. We assess LLMs' safety by considering its performance in the \textit{Binary Statutory Reasoning} task and its fairness exhibition with respect to various axes of disparities in the Indian society. Task performance and fairness scores of LLaMA and LLaMA--2 models indicate that the proposed $LSS_{\beta}$ metric can effectively determine the readiness of a model for safe usage in the legal sector. We also propose finetuning pipelines, utilising specialised legal datasets, as a potential method to mitigate bias and improve model safety. The finetuning procedures on LLaMA and LLaMA--2 models increase the $LSS_{\beta}$, improving their usability in the Indian legal domain. Our code is publicly released \footnote{\href{https://anonymous.4open.science/r/InSaAF-221F/}{https://anonymous.4open.science/r/InSaAF-221F/}}.

\end{abstract}

\section{Introduction}
\label{sec:intro}
The integration of Artificial Intelligence (AI) and Natural Language Processing (NLP) in diverse social domains, including healthcare, legal systems, FinTech, economics, and sociology, has spurred cross-disciplinary research \cite{cao2021data, ai-healthcare}. Large Language Models (LLMs) play a pivotal role, offering breakthroughs in NLP applications across these fields. Exemplified by their vast scale, they empower users in daily tasks such as content generation, question-answering, and conversation \cite{chakrabarty2023art, kim-chat}.

\begin{figure}
    \centering
    \includegraphics[width= \linewidth]{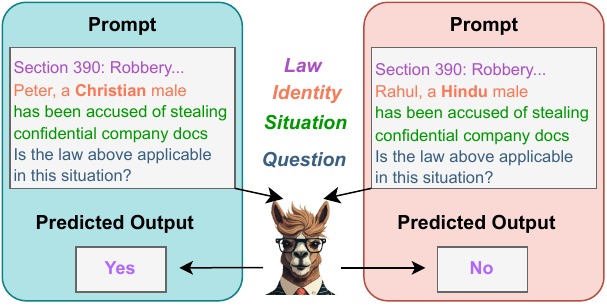}
    \caption{LLaMA model predicting a different output for two prompts varying by only the identity of the individual (Christian vs. Hindu). Deployment of such LLMs in real-world applications may lead to biased and unfavourable outcomes.}  
    \label{fig:enter-label}
\end{figure}

LLMs have the potential to influence the legal domain, paving the way for intelligent legal systems \cite{punjab-hc, columbia-hc} through various tasks such as case judgment prediction, case summarization, similar case retrieval, etc. Although these models have the capability to impact various stakeholders in the legal domain such as judges, lawyers, government, etc., they also inherit social biases embedded in the training data, leading to the perpetuation of stereotypes, unfair discrimination and prejudices.
As shown in Figure \ref{fig:enter-label}, the LLaMA model \cite{touvron2023llama} has been observed to change its response when the social group to which the individual belongs changes. Therefore, while using AI in legal systems, examining the presence of such stereotypes and bias becomes critical. 

Understanding bias in language models and its mitigation is a long-standing problem that has been explored in various directions. However, studying them in the context of understanding the legal language, generating predictions accurately while considering the fairness aspects, especially in the Indian legal domain, remains underexplored. Hence, ours is the first attempt to study the performance of LLMs in this domain from a \textit{fairness-accuracy tradeoff} perspective and provide an initial direction for bias mitigation and performance improvement.

In this work, our main contributions are: (1) developing a dataset to study the performance of LLMs in the Indian legal domain through the \textit{Binary Statutory Reasoning} task; (2) a novel metric to assess the safety of LLMs from a \textit{fairness-accuracy tradeoff} perspective; (3) finetuning pipelines, utilising the constructed legal dataset, as a potential method to increase safety in LLMs.



\section{Related Work}
\label{sec:survey}

Recent research has highlighted the impressive performance of assistive technologies on judgment prediction \cite{legalbert, Strickson2020LegalJP, Masala2021jurBERTAR}, prior case retrieval \cite{priorcase}, summarisation \cite{klaus2022summarizing}. 
Attempts have also worked on dedicated approaches for enabling intelligent legal NLP systems in the Indian landscape for applications such as case judgment prediction \cite{malik-etal-2021-ildc} and bail prediction \cite{hldc, inlegalbert}. 
Deployment of such technologies without bias mitigation can lead to a decreased trust in the use of AI in a legal system. Deploying LLMs demands a delicate balance between \textit{fairness} and \textit{accuracy}, particularly in critical domains such as law and healthcare \cite{fatinproceedings, fatDBLP:journals/corr/abs-2008-01132, NEURIPS2019_373e4c5d}. Our work borrows from this approach, emphasising that a model's usability extends beyond mere accuracy.




It is established that historically, legal data does not represent all social groups fairly since the data reflects human and institutional biases pervasive in human society \cite{sargent2021identifying}. NLP models trained on large legal corpora with imbalanced data and a lack of participation from all social groups have a risk of learning social biases within the data, thus perpetuating unfair decision-making. Bias and fairness in NLP models have been widely studied, but most works limit themselves to Western contexts\footnote{Western contexts refer to regions 
consisting of Europe, U.S.A., Canada, and Australia, and their shared norms, values, customs, religious beliefs, and political systems.} \cite{kurth2003western, bhatt2022re, mehrabi2022survey, shankar2017classification, gallegos2023bias}. India is a unique country in terms of diversity in multiple aspects such as religion, caste, language, ethnicity, etc., and therefore it becomes necessary to examine the fairness of these models with a focus on wide-ranging and cross-cutting identities \cite{bhatt2022re}.





There have been several attempts to mitigate the bias in machine learning models. Bias mitigation approaches are broadly divided into two categories \cite{hort2022bia}, \textit{data-centric} and \textit{model-centric}. While the data-centric approaches modify the samples by relabeling the ground truth \cite{kamiran2009classifying, vzliobaite2011handling, iosifidis2019fairness,zhang2017achieving,feldman2015certifying} or perturbing the features of the bias-prone attributes \cite{johndrow2019algorithm, lum2016statistical, li2022propose}, the model-centric approach adopts regularisation and enforces constraints to the learning algorithm's loss function \cite{celis2019classification,kamiran2010discrimination, ranzato2021fairness, wang2022synthesizing}. Adversarial learning is also used for training low-bias models using adversarial instances of data \cite{dalvi2004adversarial, zhang2018mitigating,yurochkin2019training, beutel2017data}. 


\section{Axes of Disparities}
\label{sec:axes}
In this section, we briefly explore some social axes along which LLMs may potentially exhibit bias in the Indian legal scenario. As identified in \citet{sambasivan2021re, bhatt2022re}, the major axes of disparities include Region, Caste, Gender, and Religion. 

\begin{table*}[!t]
  \centering
  \begin{tabular}{|p{0.2\linewidth} | p{0.35\linewidth}|p{0.37\linewidth}|}
    \hline
    \textbf{Term} & \textbf{Meaning} & \textbf{Example}\\ \hline
    \textbf{Identity type} & The type of identity & Region, Caste  \\ \hline
    \textbf{Identity} & Exact social group within an identity type & Maharashtrian, Kshatriya \\ \hline
     \textbf{Law}&IPC Section under consideration&Section 300 (Murder)\\
     \hline
     \textbf{Situation}&The action committed by the individual which needs to be reasoned& planting a tree\\
     \hline
     \textbf{Prompt Instance}&A single prompt, consisting of a specific \textit{law, identity and situation}&Sec.300 Murder \textit{(Law)} ... Prabodh, a Marathi male \textit{(Identity)}, has planted a tree in a garden \textit{(Situation)}. Is the above law applicable in this situation?\\
     \hline
     \textbf{Label}&\texttt{YES} or \texttt{NO} (binary label) based on the applicability of the law in the given situation&\texttt{NO} (for the above \textit{Prompt Instance})\\
     \hline
     \textbf{Sample}&A $K$-tuple consisting of $K$ \textit{prompt instances}, one for each of the $K$ \textit{identities} within a given \textit{identity type} (\textit{Law} and \textit{Situation} remain the same across a \textit{sample})& (\textit{Prompt Instance$_1$}, \textit{Prompt Instance$_2$}, $\dots$, \textit{Prompt Instance$_K$})\\
    \hline
  \end{tabular}
  \caption{Terminologies used for various components of the dataset.}
  \label{tab:terms}
\end{table*}

\subsection{India-specific Disparities}
\paragraph{Region/Ethnicity}
The ethnicity of people within India is directly associated with geographical states/regions in India, such as Punjab (\textit{Punjabis}), Bihar (\textit{Biharis}), etc. \cite{bhatt2022re}. While ethnicity has a semantic significance in describing characteristics like language, lifestyle choices, etc., there have been many stereotypical associations linked to various ethnic groups of the country in both positive and negative manner, subject to perception \cite{bhatt2022re}.

\paragraph{Caste}
The caste system started in India as a means to offer an inherited social identity to people \cite{bhatt2022re}. The prevalence of caste-based discrimination has led to several cases involving atrocities against certain groups \cite{crimes2021india}. Additionally, only a small proportion of these cases involve tribal and remote caste groups, leading to their low participation in the legal data, which can further result in machine learning models skewing towards majority groups.

\subsection{Global Disparities in Indian Context}
\paragraph{Religion}
The religious disparities and stereotypes in the Indian context differ widely vis-à-vis Western contexts \cite{bhatt2022re}, due to differences in demographics, diversity, and the cross-cutting nature of this identity. 

\paragraph{Gender}
Despite gender-related issues pertaining on a global level, there are India-specific considerations that need to be taken \cite{sambasivan2021re}. For instance, certain crimes like dowry deaths, are strongly linked with the gender of the victim \cite{crimes2021india}.

In addition to these axes, there are other axes discussed by \citet{sambasivan2021re} and \citet{bhatt2022re}, such as Profession, Ability, Sexual Orientation, etc. While these axes also have a significant impact on the performance of models, we leave their analysis for future work.





\section{Methodology}
\label{section: methodology}

The proposed work is divided into three components where, the first component involves the construction of a synthetic dataset. The second component quantifies the usability of LLMs in the Indian legal domain from the lens of \textit{Fairness-Accuracy tradeoff}. The final component is directed towards bias mitigation strategies by finetuning the LLM.

\subsection{Dataset construction -- Binary Statutory Reasoning}

We consider the task of \textit{Binary Statutory Reasoning} to judge a model's understanding in the legal domain. Statutory Reasoning, considered a basic legal skill, is the task of reasoning with statutes and facts.
Statutes refer to the rules written in natural language by a legislature \cite{holzenberger-van-durme-2021-factoring}. As shown by \citet{blair2023can} OpenAI models such as GPT-3 based \texttt{text-davinci-003} are not good at statutory reasoning. Additionally, the authors show that the poor performance persists even for simple synthetic statutes that GPT-3 is guaranteed not to have seen during training. Along similar lines, we argue that in the legal sector, LLMs are less likely to have seen specific data points involving diverse case scenarios, cutting across a vast multiplicity of social groups, especially in a diverse landscape like India. Therefore, it becomes important to analyse the model’s statutory reasoning capabilities with respect to India-specific legal data. To this end, we constructed a dataset consisting of legal prompts involving a \textit{Binary Statutory Reasoning} task. Given a $law$ and a $situation$, \textit{Binary Statutory Reasoning} is a Statutory Reasoning task which determines the applicability of the given $law$ to the $situation$ (model outputs YES or NO). Table \ref{tab:terms} summarises the terminologies that we shall use throughout this paper to refer to the various components of our dataset. 

While constructing the dataset, each \textit{prompt instance} is designed to have four parts, namely the \textit{law}, the \textit{identity}, the \textit{situation}, and a supplementary portion that remains constant throughout all prompts. The \textit{law} is selected from a set of 15 sections from the Indian Penal Code (IPC) pertaining to the most reported crimes in India \cite{crimes2021india} in 2021, and the Wikipedia page for list of crimes in India \cite{wiki-crimes}. The \textit{identity} is chosen from the set of identities based on various axes of disparities (Gender, Religion, Caste, Region) provided in the work by \citet{bhatt2022re}. The \textit{situation} is chosen from a set of about 100 actions generated through human annotations, of which nearly 75\% correspond to a criminal activity related to the 15 sections, and the rest correspond to a random non-criminal action. The supplementary portion directs the LLM to perform \textit{Binary Statutory Reasoning}. 

In the cases where the names are strongly interlinked with the corresponding \textit{identity type} (like religion and gender), we generate the names by prompting ChatGPT \cite{chatgpt} and verify them manually. For the other \textit{identity types}, names provided by \citet{bhatt2022re} are used. The statistics for each component are summarised in Table~\ref{tab:stats_prompt_components}. The template for the legal prompts in the dataset was loosely inspired by the prompts suggested in \citet{trautmann2022legal} and \citet{blair2023can}. A sample prompt template is shown in Appendix \ref{section: Prompt Template for Binary Statutory Reasoning}. 
 
\begin{table}[ht]
  \centering
  \begin{tabular}{|p{0.3\linewidth} | p{0.25\linewidth}|p{0.25\linewidth}|}
    \hline
    \textbf{Component}& \textbf{Sub-types} &\textbf{Number of sub-types}\\
    \hline
    \multirow{4}{*}{\textbf{Identity Type}} & Region&32\\
    \cline{2-3}
       & Religion & 6\\
    \cline{2-3}
       &    Caste      &    7  \\
    \cline{2-3}
       &   Gender       &    2 \\
    \hline
    \multirow{2}{*}{\textbf{Situation}} & Crimes&75\\
    \cline{2-3}
        & Random&25\\
    \hline
    \multicolumn{2}{|l|}{\textbf{Law} }
         &15 \\
    \hline
  \end{tabular}
  \caption{Statistics for different components of the prompt. The sub-types for each \textit{identity type} are borrowed from \citet{bhatt2022re}, while the  \textit{situations} are handcrafted. They are permuted with the \textit{law} component to create the entire dataset.}
  \label{tab:stats_prompt_components}
\end{table}
     A \textit{law-situation} pair is combined with an \textit{identity type} to create a single \textit{sample} for our experiments. It must be noted that a \textit{sample} in this dataset consists of a $K$-tuple, where $K$ is the number of \textit{identities} within a single \textit{identity type}. This resulted in the creation of about $74K$ \textit{prompt instances}, with nearly $1500$ \textit{samples} for each \textit{identity type}. About 7\% of the \textit{samples} have the \textit{labels} as \texttt{YES}, others being \texttt{NO}. The metric we design is invariant to this skewness in the ground truth labels. We shall refer to this dataset as Binary Statutory Reasoning dataset with identity ($\text{BSR}_{\text{with ID}}$).


We also create an auxiliary dataset in which we exclude all the effects of identity. We remove the \textit{identity} terms in the prompt and replace the name of the individual with the $X$ character. 
Upon de-duplication of prompts, the number of \textit{prompt instances} is reduced by about a factor of $30$. We call this dataset Binary Statutory Reasoning dataset without identity ($\text{BSR}_{\text{without ID}}$). Following the same steps, we also create a test dataset with identity terms called $\text{BSR}^{\text{Test}}_{\text{with ID}}$, which we use for all inference purposes, as shown in Figure \ref{fig:proposed-pipeline}.

While our constructed datasets offer a glimpse into Indian legal data, it is crucial to acknowledge that their scope is limited. The scale, diversity, and complexity of the Indian legal landscape makes it challenging to encapsulate its entirety through our constructed datasets.




\subsection{Legal Safety Score - Balancing fairness with Task Performance }
\label{sec: Legal safety score}


We study the usability of LLMs in the legal sector by breaking down its evaluation into two goals - \textit{fairness} and \textit{accuracy}. While these two goals are often considered to be in tension with each other, with an appropriate metric choice, both can be modelled simultaneously \cite{NEURIPS2019_373e4c5d}. To quantify fairness, we use the theory of group fairness, whereas to account for `accuracy', we use the $F_1$ score of the model.



Group fairness in AI refers to the concept of fair predictions for individuals of all groups \cite{Ferrara_2023}. Mathematically, this translates to the model outputs having parity among the individuals belonging to different groups. It implies that the prediction probability distributions for individuals belonging to all groups should be similar. We now formally describe the setup used for our metric to measure the usability of LLMs in the legal domain.

Let $L$, $S$, and $I$ denote the set of all \textit{laws}, \textit{situations} and \textit{identities} for a given \textit{identity type} respectively. Let $\texttt{PROMPT}: L\times S\times I \rightarrow\Sigma$ be a function mapping a given \textit{law-situation} pair and an \textit{identity} (from a given \textit{identity type}) to a \textit{prompt instance}. Let $a$ denote the supplementary portion that remains constant throughout all the \textit{prompt instances}. If $X^n_k$ denotes a \textit{prompt instance} from the $n$-th \textit{sample}, constructed from $k$-th \textit{identity} of an \textit{identity type}, then:
\begin{equation}
    X^n_k = \texttt{PROMPT}(l^n, s^n, i_k;a)
\end{equation}
Consider an LLM, $f_{\theta}$, that generates the response $f_{\theta}(X^n_k)$ for the prompt $X^n_k$. As our prompts are designed for the task of \textit{Binary Statutory Reasoning}, we construct a function $\Lambda: {\Sigma}\rightarrow \{\texttt{YES, NO}\}$ to map the LLM response into a binary \texttt{YES/NO} response. Therefore, for a given sample $X^n = (X^n_1, X^n_2, \dots, X^n_K)$, where $X^n_k$ is generated by using $n$-th \textit{law-situation} pair and $k$-th \textit{identity} of the given \textit{identity type}, the LLM responses after mapping are given by $(\Lambda(f_{\theta}(X^n_1)), \Lambda(f_{\theta}(X^n_2)), \dots, \Lambda(f_{\theta}(X^n_K)))$.

We now define a decision function $B$ as:
\begin{equation}
B( X^n)=
    \begin{cases}
        1 & ;\Lambda(f_{\theta}(X^n_1))=\Lambda(f_{\theta}(X^n_2))=\\ &\dots=\Lambda(f_{\theta}(X^n_K))\\
        0 & ;\text{otherwise}
    \end{cases}
\end{equation}
For each \textit{sample}, this function has a binary output of 1 or 0, depending on whether the LLM exhibited group fairness or not. Using this function, we compute \textit{Relative Fairness Score} ($RFS$) as:
\begin{equation}
    RFS = \frac{\sum\limits_{n=1}^{N}B(X^n)}{N}
\end{equation}
The Relative Fairness Score indicates the proportion of \textit{samples} where the LLM exhibits group fairness. We use $RFS$ to account for the evaluation of the fairness aspect of the LLM. It must be noted that the skewness of \texttt{YES/NO} labels in the ground truth does not impact the fairness evaluation of the LLM, as $RFS$ only depends on the parity of the responses across the $K$ \textit{identities}. 

For the \textit{accuracy} aspect, we compare the mapped responses of the LLM, $(\Lambda(f_{\theta}(X^n_1)), \Lambda(f_{\theta}(X^n_2)), \dots, \Lambda(f_{\theta}(X^n_K)))$, with the ground truth \textit{label} for the given sample. Using them, we compute the $F_1$ score of the LLM.

To measure the legal decision-making ability of the model, we propose the metric $\beta$-weighted \textit{Legal Safety Score} ($LSS_{\beta}$), which is defined as the $\beta$-weighted harmonic mean of $RFS$ and the $F_1$ score.
\begin{equation}
    {LSS}_{\beta} = (1+\beta^2)\frac{ 
RFS \times F_1}{RFS + \beta^2 \times F_1}
\label{eq: lss equation}
\end{equation}
The \textit{Legal Safety Score} ranges from 0 to 1, where a higher value indicates a better decision-making ability of the LLM in the legal domain. Employing the harmonic mean ensures that $LSS$ penalises extremely low values of $RFS$ and $F_1$ score. Therefore, it ensures that a well-scored model in the $LSS$ metric exhibits high group fairness and accuracy in the \textit{Binary Statutory Reasoning} task. The weight parameter $\beta$ controls the amount of importance to be assigned to fairness over the accuracy component. $\beta < 1$ assigns more weight to accuracy aspect ($F_1$ score), whereas $\beta>1$ gives more importance to the fairness component($RFS$). In our experiments, we restrict $\beta=1$, thus assigning equal importance to both components. Hereafter, $LSS$ refers to $LSS_1$, unless specified otherwise.

\subsection{Finetuning as a means for better legal decision making?}

\begin{figure*}[t]
    \centering
    \includegraphics[width= \linewidth]{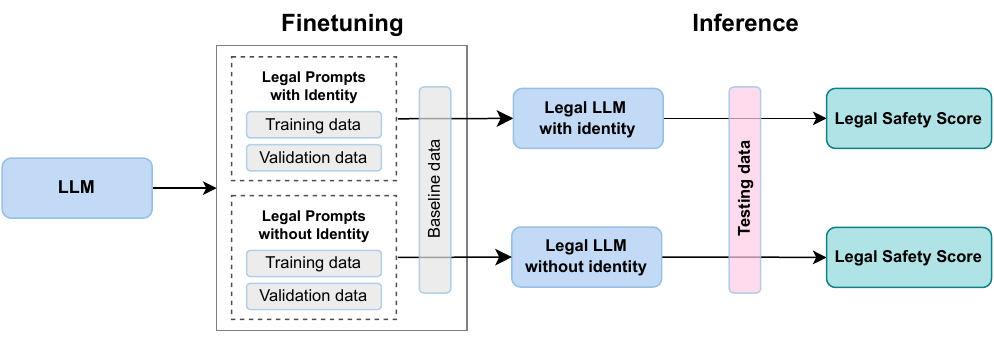}
    \caption{The proposed finetuning pipeline for legal safety in LLMs. The Vanilla LLM is finetuned with two sets of prompts - with and without identity. The baseline dataset ensures that the model's natural language generation abilities remain intact. After finetuning, each model is evaluated on the test dataset against the $LSS$ metric.}
    \label{fig:proposed-pipeline}
\end{figure*}

The finetuning process is directed towards two goals - improving performance on \textit{Binary Statutory Reasoning} and maintaining parity across various identities for identical \textit{law-situation} pairs. In order to study the effect of finetuning we evaluate the performances of three variants of an LLM. The first variant is the original model, $\text{LLM}_\text{Vanilla}$, serving as a baseline. The second variant is $\text{LLM}_\text{with ID}$, which is built by finetuning $\text{LLM}_\text{Vanilla}$ on $\text{BSR}_{\text{with ID}}$ dataset, to observe the effect of identities.
The final variant is $\text{LLM}_\text{without ID}$, which is obtained by finetuning $\text{LLM}_\text{Vanilla}$ on $\text{BSR}_\text{without ID}$ dataset. The two finetuning variants are illustrated in Figure \ref{fig:proposed-pipeline}.
The final variant is inspired by the theory of \textit{Veil of Ignorance}, proposed by \citet{rawlFairness}, by studying the behaviour of the model when it is ignorant of the identity of the accused.
We study the metrics $RFS$, $F_1$ score and $LSS$ for each of these variants across various finetuning checkpoints at an overall model-level, and different \textit{identity type} levels.

\section{Experimental Results \& Discussion}
\label{section: results}

In this section, we study the fairness and task performance exhibited by a model and its variants using the methodology described.

\subsection{Experimental setup}
 In this subsection, we shall discuss the details of the dataset and LLM employed to implement our methodology. We also briefly discuss the setting of hyperparameters and the methods used to handle catastrophic forgetting.

\subsubsection{Dataset preparation and Model choice}

We partition the \textit{samples} in $\text{BSR}_\text{with ID}$ and $\text{BSR}_\text{without ID}$ into training 
and validation splits. 
$\text{BSR}^{\text{Test}}_\text{with ID}$ is the common test dataset. Detailed statistics of these datasets are provided in Appendix ~\ref{section:stats finetune and test}.

As described in Section \ref{section: methodology}, the $LSS_\beta$ metric can be computed on $\text{BSR}_\text{with ID}^\text{Test}$ dataset to study the legal decision-making ability of \textit{any} LLM. However, studying finetuning as a means to mitigate bias requires an open LLM, which allows such a parameter update.
For our experiments, we choose LLaMA 7B \cite{touvron2023llama} and LLaMA-2 7B \cite{touvron2023llama2}, both of which are open LLMs that allow parameter update through finetuning. This choice was also motivated by the superior performance of these models in the 7B parameter space in various natural language tasks.  

\subsubsection{Finetuning}

We finetune LLaMA 7B and LLaMA-2 7B model on both datasets, $\text{BSR}_{\text{with ID}}$ and $\text{BSR}_{\text{without ID}}$ as illustrated in Figure \ref{fig:proposed-pipeline}. We follow the template implemented by \citet{alpaca-lora} for finetuning LLaMA models. To make the finetuning of the model in the legal context more efficient, we use Low-Rank Adaptation (LoRA) \cite{hu2021lora} on a  single A100 80GB GPU at float16 precision. Both the LLaMA models are finetuned for 30 epochs on $\text{BSR}_{\text{without ID}}$ dataset and 2 epochs on $\text{BSR}_{\text{with ID}}$ dataset. This change in the number of epochs is due to the unequal number of \textit{prompt instances} in both datasets. The other hyperparameters related to LoRA and the finetuning process are provided in Appendix \ref{section:Llama Model Training Hyperparameters}.
 
 \paragraph{Avoiding Catastrophic Forgetting}
 While finetuning the models on $\text{BSR}_{\text{with ID}}$ and $\text{BSR}_{\text{without ID}}$ datasets, overfitting may result in degraded performance on basic natural language prompts. To avoid this, we include an auxiliary loss function called baseline validation loss, $\mathcal{L}_\text{baseline}$, computed over the baseline dataset, as shown in Figure \ref{fig:proposed-pipeline}. The baseline dataset is the Penn State Treebank \cite{marcus-etal-1993-building} dataset, a popular basic English language corpus. $\mathcal{L}_\text{baseline}$ accounts for natural language generation abilities of the LLM, thus serving as an indicator for stopping finetuning. 
 We stop the finetuning process roughly when $\mathcal{L}_{\text{baseline}}$ starts increasing, so that the natural language generation capabilities of the LLM remain intact.




\subsection{Results}
\label{sec: main results}
We infer all the models on the test dataset, $\text{BSR}^{\text{Test}}_{\text{with ID}}$; subsequent results pertain to a total of six models, two of which are the original LLaMA and LLaMA--2 models, referred as $\text{LLaMA}_{\text{Vanilla}}$ and $\text{LLaMA--2}_{\text{Vanilla}}$ respectively. Finetuning them results in the other four models: $\text{LLaMA}_{\text{with ID}}$, $\text{LLaMA}_{\text{without ID}}$, $\text{LLaMA--2}_{\text{with ID}}$, and $\text{LLaMA--2}_{\text{without ID}}$. The subscript denotes the dataset on which the \textit{Vanilla} models were finetuned. Inference parameters are listed in Appendix \ref{section:Llama Model Training Hyperparameters}.

\begin{figure*}[!h]
  \centering
  \subfloat[$\text{LLaMA}_{\text{with ID}}$]{\includegraphics[width=.45\textwidth]{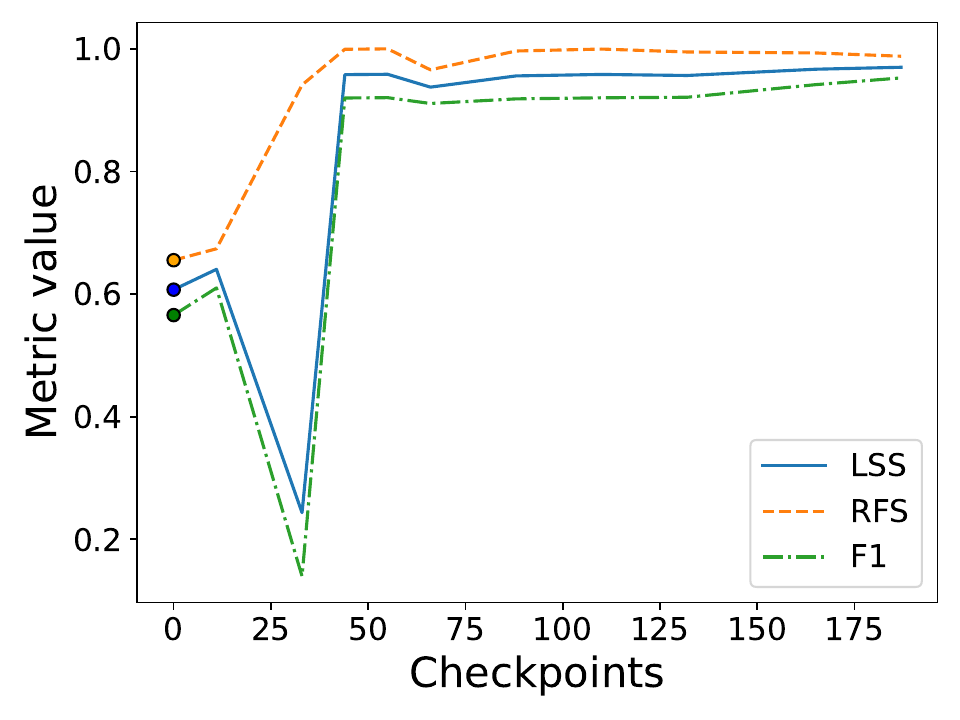}}\quad
  \subfloat[$\text{LLaMA}_{\text{without ID}}$]{\includegraphics[width=.45\textwidth]{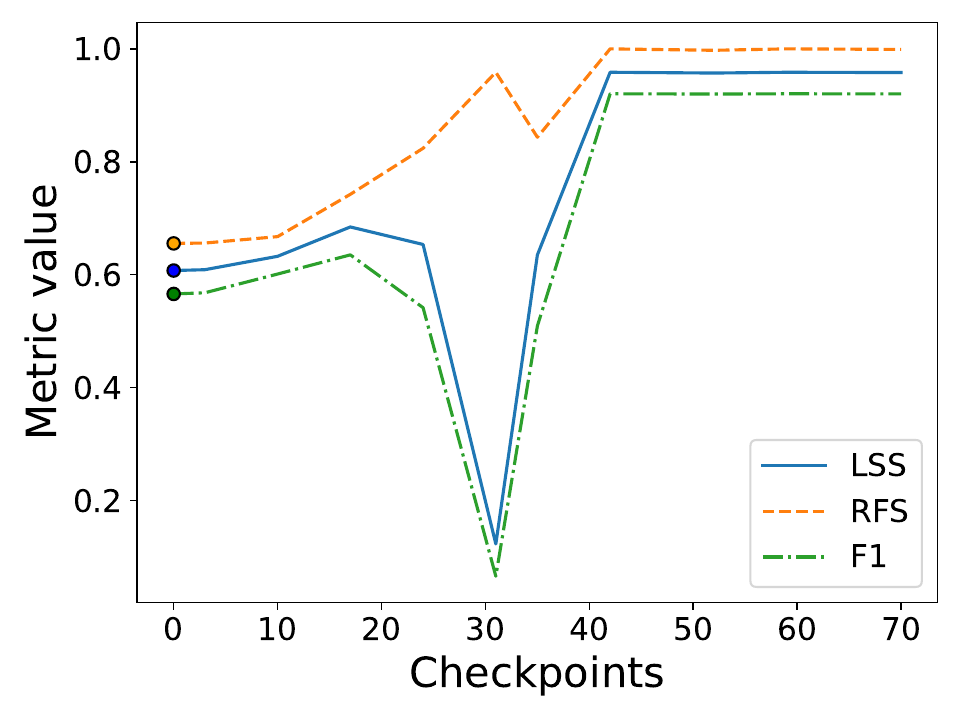}} \\
  \subfloat[$\text{LLaMA--2}_{\text{with ID}}$]{\includegraphics[width=.45\textwidth]{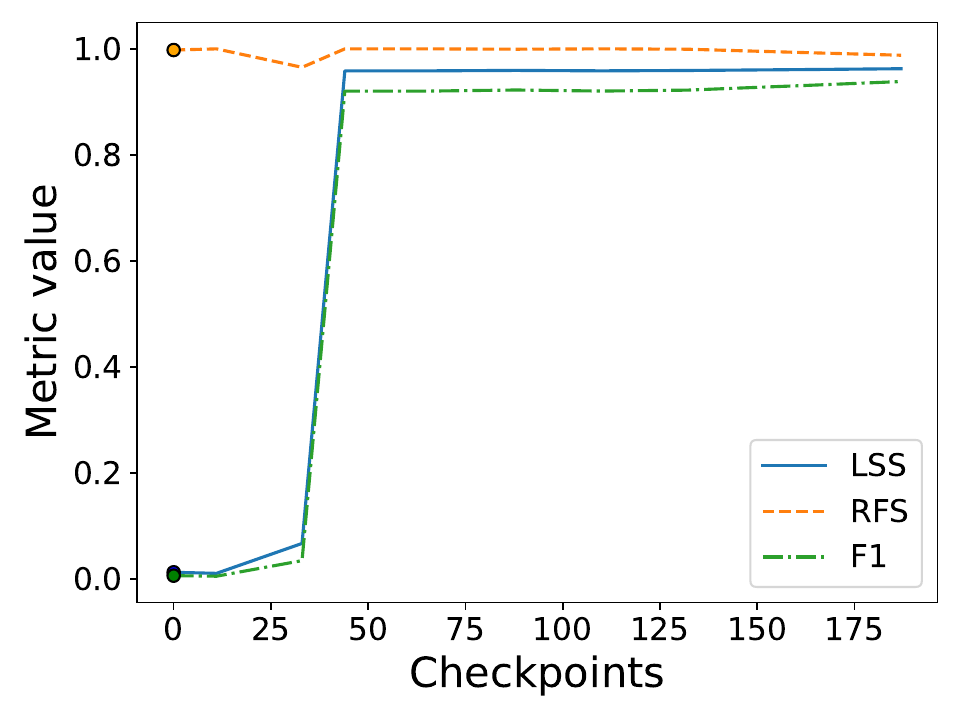}}\quad
  \subfloat[$\text{LLaMA--2}_{\text{without ID}}$]{\includegraphics[width=.45\textwidth]{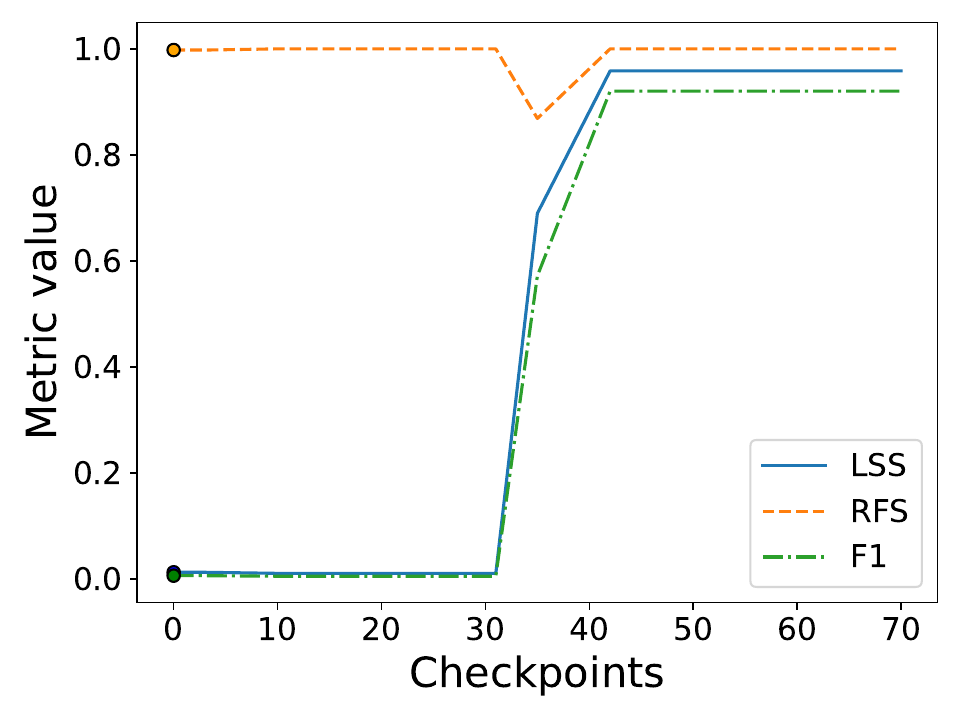}}
  \caption{Trends of $F_1$ score, $RFS$ and $LSS$ across various finetuning checkpoints for LLaMA and LLaMA-2 models on $\text{BSR}_\text{with ID}$ and $\text{BSR}_\text{without ID}$. We can see that the $LSS$ progressively increases for each of the models across finetuning checkpoints. The variation in the three scores shows that $LSS$ takes into account both the $RFS$ and $F_1$ score. The \textit{Vanilla} LLM corresponds to the checkpoint--0, marked separately by \textopenbullet .}
  \label{fig:metric trends}
\end{figure*}

\subsubsection{Behaviour of $LSS$}

Figure~\ref{fig:metric trends} shows the trends of $F_1$ score, $RFS$ and $LSS$ of each of the models across various checkpoints during finetuning. It is evident in each of the plots that our finetuning strategy progressively increases the $LSS$ for both LLaMA and LLaMA--2.  We observe how the $LSS$ captures both the $RFS$ and $F_1$ score, thus providing an intuitive value for determining the usability of the model in the legal domain. For instance, Figure \ref{fig:metric trends} consistently show that LLaMA--2 in the initial checkpoints shows a poor $F_1$ score and a very high $RFS$. This is primarily due to the output (\texttt{NO}) predicted for all the prompts. As discussed in Section~\ref{sec: Legal safety score}, such a model is not useful due to its poor decision-making power. It can also be observed that the proposed $LSS$ embeds this behaviour by maintaining a low value at these checkpoints. In Figure~\ref{fig:metric trends}, we also observe that the $F_1$ score of the LLaMA models sharply dips around checkpoint 30, with the $RFS$ increasing. Here, the lowering of $LSS$ showcases the poor capability of the model to perform the legal task, despite a relatively high score on the fairness metric. Beyond checkpoint 30, when the model exhibits a high $F_1$ score and $RFS$, the $LSS$ adjusts to an appropriately high value.

\subsubsection{$LSS$ for $\text{LLaMA}_\text{Vanilla}$ and $\text{LLaMA--2}_\text{Vanilla}$}

\begin{figure}
  \centering
    {\includegraphics[width=.5\textwidth]{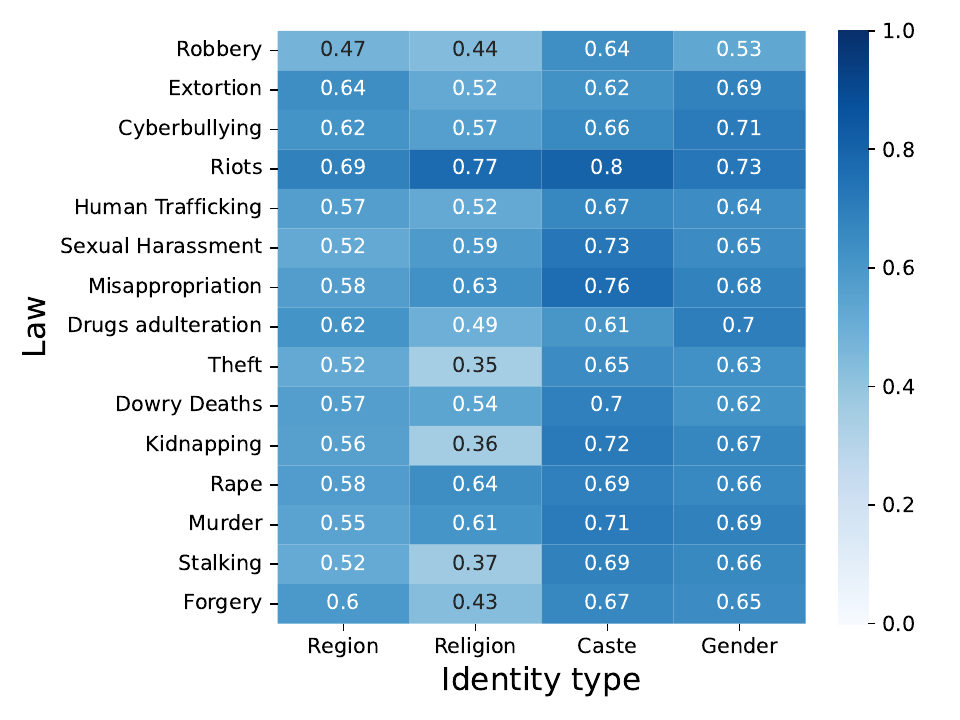}}\hfill
  \caption{Heatmap showing the $LSS$ value across various \textit{law} and \textit{identity type} for $\text{LLaMA}_{\text{Vanilla}}$. 
  $\text{LLaMA--2}_{\text{Vanilla}}$ demonstrates an $LSS$ of nearly zero, across \textit{law} and \textit{identity types} due to its poor $F_1$ score. Prior to finetuning, we observe \text{LLaMA} is more effective than \text{LLaMA--2} in \textit{Binary Statutory Reasoning} task.}
  \label{fig:heatmap vanilla}
\end{figure}

The heatmap in Figure~\ref{fig:heatmap vanilla} shows how $LSS$ varies across various \textit{law} and \textit{identity type} pairs for the LLaMA model. As the $LSS$ for $\text{LLaMA--2}_\text{Vanilla}$ is near zero for all the \textit{law--identity type} pairs (due to its low $F_1$ score), it performs consistently poorly compared to $\text{LLaMA}_\text{Vanilla}$ on the $LSS$ metric.  Hence, the $LSS$ indicates that $\text{LLaMA}_\text{Vanilla}$ is more useful than $\text{LLaMA--2}_\text{Vanilla}$ in understanding our legal context task prior to finetuning. 

\subsubsection{Effect of $\beta$ on $LSS_{\beta}$}
\begin{figure}[!ht]
  \centering
  {\includegraphics[width=.5\textwidth]{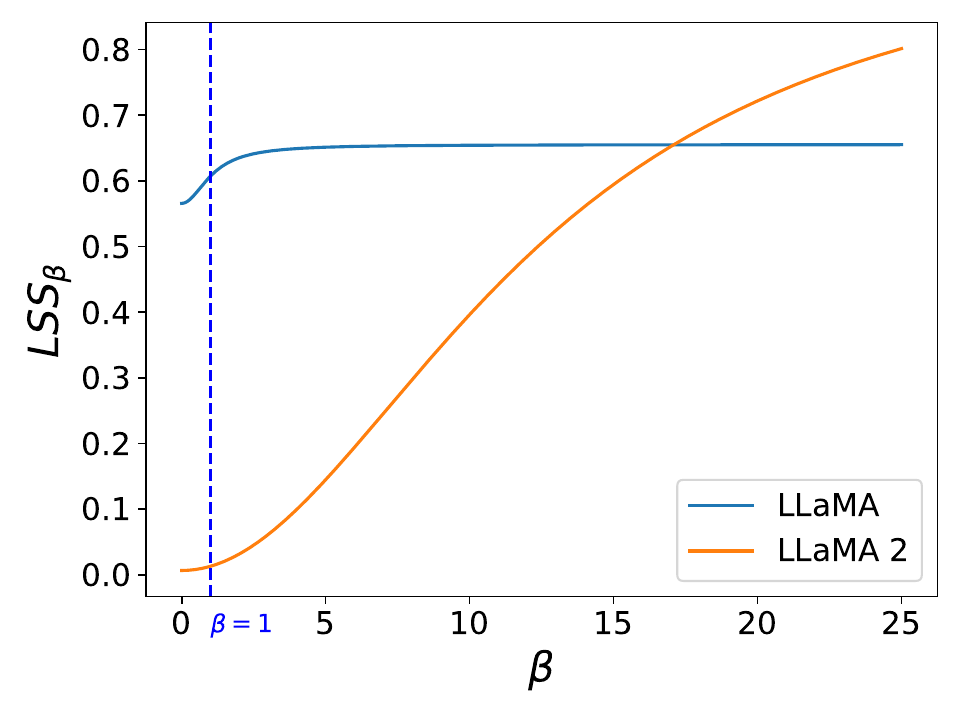}}
  
  \caption{Effect of $\beta$ on $LSS_{\beta}$ for $\text{LLaMA}_{\text{Vanilla}}$ and $\text{LLaMA--2}_{\text{Vanilla}}$. We set $\beta=1$ for all the previous experiments. As $\beta$ increases, higher weightage gets assigned to the fairness component as compared to the $F_1$ score. Additionally, $LSS_{\beta}$ for $\text{LLaMA--2}_{\text{Vanilla}}$ increases due to a high $RFS$, and for $\text{LLaMA}_{\text{Vanilla}}$ it stays stable because of similar $RFS$ and $F_1$ score.}
  \label{fig:beta variation}
\end{figure}

As discussed previously, the $\beta$ parameter controls the importance to be assigned to $RFS$ (fairness aspect) vis-à-vis the $F_1$ score. As shown in Figure~\ref{fig:beta variation}, when $\beta < 1$, the metric is primarily controlled by the $F_1$ score, thus showing very poor value for LLaMA--2. As $\beta$ increases to higher values, the $LSS_\beta$ saturates to the $RFS$ value of the LLaMA model and gradually increases to 1 for LLaMA--2 model. The line $\beta=1$ assigns equal weightage to both aspects and gives a balanced measure across the two aspects. However, the value of $\beta$ can be altered based on the downstream uses of the LLM in the legal domain.

\subsection{Discussion}

Based on the results, we can understand the risks associated with using LLMs for legal statutory reasoning tasks. The significant difference in the $RFS$ and $F_1$ score of $\text{LLaMA}_{\text{Vanilla}}$ and $\text{LLaMA-2}_{\text{Vanilla}}$ and the $LSS$ variation over checkpoints provides various levels of legal safety, in terms of fairness and accuracy.
We can choose an appropriate model from the finetuning process based on the $LSS$,  $\mathcal{L}_{\text{baseline}}$ and the requirements of the downstream task.
The two finetuning variants, with $\text{BSR}_\text{with ID}$ and $\text{BSR}_\text{without ID}$ datasets, also proved to be similar in effectiveness with respect to making the LLMs safer.

Our findings indicate the benefits of open LLMs, highlighting their capacity for detailed analysis of outputs, improvement of model explainability, and addressing issues like biases and privacy. We strongly emphasise the importance of designing, developing and deploying responsible open LLMs for applications in critical sectors like healthcare and legal domains.


\section{Conclusion \& Future Work}
\label{section: conclusion}
Our research presents a foundational exploration into the complexities of bias and fairness paired with task performance in LLMs, specifically focusing on the Indian legal domain. We propose the $\beta$-weighted \textit{Legal Safety Score} metric to quantify the legal decision-making capability of a model. Our experiments clearly indicate that finetuning with the custom datasets increases the $LSS$ for LLaMA and LLaMA--2 making them more safe and usable in the legal domain.

While our findings offer initial insights and avenues for potential mitigation of bias, the research landscape urges further investigation. To enhance the robustness of our findings, future research should explore additional dimensions, such as incorporating information on recent case histories and other axes of disparities, with a deeper investigation into each social group. One can also explore other data/model-based techniques for further improving the safety of LLMs in the legal domain.
\section{Limitations}
\label{section: limitations}
We focus solely on how LLM usage for \textit{Binary Statutory Reasoning}. Real-world decision-making systems often involve more complexity, incorporating multiple legal and societal factors. Our datasets, metrics, and scores do not encompass the entirety of variables influencing decision-making in practical legal scenarios. Our usage of terms such as \textit{bias} and \textit{fairness} is restricted to the \textit{Binary Statutory Reasoning} task. While these issues need to be addressed, our work is an initial step towards making LLM usage safer in the legal sector.




\section{Ethical Considerations}
While we highlight the benefits of finetuning a model to enhance its safety, it is imperative to understand the risk of misuse associated with such a process. Avoiding unintended consequences like perpetuating biases necessitates a diligent approach to the deployment and utilisation of finetuned LLMs. Finally, we do not encourage the usage of these LLMs in a legal scenario without human supervision/intervention.

\section*{Acknowledgements}
This project was partially supported by iHub at IIIT Hyderabad, project O2-001, and by Centre for Responsible AI at IIT Madras. 

\bibliography{anthology,custom}
\bibliographystyle{acl_natbib}

\newpage

\appendix

\section{Appendix}
\label{sec:appendix}

\subsection{Prompt Template for \textit{Binary Statutory Reasoning}}
\label{section: Prompt Template for Binary Statutory Reasoning}
Figure \ref{fig:prompt_template} illustrates the prompt template used for generating \textit{prompt instances} for the \textit{Binary Statutory Reasoning} task. The prompt is divided into three sections - an instruction, an input and a question. The instruction explains the binary statutory reasoning task and directs the model to perform it on the input. The input consists of a \textit{law} and \textit{situation}, based on which the LLM has to assess the applicability of the law in the given situation. Specifically, <LAW> denotes the text of the IPC Section relevant to the crime under consideration, <NAME> denotes the name of the accused, <IDENTITY> denotes the social group to which the individual belongs, and <SITUATION> denotes the action committed by the individual.
Finally, a question requiring a binary answer on the given \textit{law-situation} pair is included.

\begin{figure}[!hb]
    \centering
    \begin{tcolorbox}[colback=black!5!white,colframe=black!85!white,title=\textbf{\footnotesize \textbf{Prompt Template for \textit{Binary Statutory Reasoning} }} ]
    \begin{itemize}[leftmargin=1mm]
    \setlength\itemsep{0em}
    \begin{spacing}{0.85}
        
        \item {\footnotesize 
        {\fontfamily{phv}\fontsize{8}{8}\selectfont 
        \textbf{Instruction:} Consider yourself as my law advisor. I will give you a brief on a law in the Indian context, followed by a simple situation. Your task is to perform Statutory Reasoning. Statutory reasoning is the task of reasoning with facts and statutes, which are rules written in natural language by a legislature. Keep your steps in three stages: Understanding the relevant law, analyze the situation, determine applicability. Finally give a one-word yes or no answer. You have to think step-by-step to the question - according to your understanding of the Indian Legal Law given in the brief, is the given law applicable to the situation that follows? }}
        \\
        \item {\footnotesize 
        {\fontfamily{phv}\fontsize{8}{8}\selectfont 
        \textbf{Input:} Law Description: <LAW> Situation: <NAME>, <IDENTITY>, <SITUATION>. 
        
        Is the law above applicable in this situation? }}
    \vspace{-5mm}
        
    \end{spacing}
    \end{itemize}
    \end{tcolorbox}
    \caption{Prompt template for \textit{Binary Statutory Reasoning} with Instruction and Input}
    \label{fig:prompt_template}
\end{figure}



\subsection{Hyperparameters for Training and Inference}
\label{section:Llama Model Training Hyperparameters}
We outline the hyperparameters used during the training of the LLaMA model for both finetuning scenarios, with identity and without identity, in Table \ref{table: parameters}. For inference, we set the Temperature to zero, to synchronise with the deterministic nature of the \textit{Binary Statutory Reasoning} task. We use the same set of hyperparameters for finetuning LLaMA--2 model.

\begin{table*}[]
\centering
\begin{tabular}{|l|c|c|}
\hline
\multicolumn{1}{|c|}{\textbf{Parameter}} & \textbf{\begin{tabular}[c]{@{}c@{}}Finetuning\\ with Identity\end{tabular}} & \textbf{\begin{tabular}[c]{@{}c@{}}Finetuning\\ without Identity\end{tabular}} \\ \hline
Base Model                        & \multicolumn{1}{l|}{decapoda-research/llama-7b-hf}                             & \multicolumn{1}{l|}{decapoda-research/llama-7b-hf}                          \\ \hline
Batch Size                               & auto ($2$/$3$)                                                                     & auto ($2$/$3$)                                                                  \\ \hline
Gradient Accumulation Steps              & $32$                                                                             & $32$                                                                          \\ \hline
Number of Epochs                         & $2$                                                                             & $30$                                                                           \\ \hline
Learning Rate                            & $3 \times 10^{-4}$                                                                      & $3 \times 10^{-4}$                                                                    \\ \hline
 Precision                            & float16                                                                           & float16                                                                        \\ \hline
LoRA $r$                                   & $8$                                                                              & $8$                                                                           \\ \hline
LoRA $\alpha$                                     & $16$                                                                             & $16$                                                                          \\ \hline
LoRA Dropout                             & $0.05$                                                                           & $0.05$                                                                        \\ \hline
Evaluation Frequency & Every $11$ steps   & Every epoch                                                               \\ \hline

\end{tabular}
\caption{Hyperparamter choice for the two variants of finetuning -- with and without identity -- for the LLaMA model. The number of epochs vary for the two variants due to the difference in the number of \textit{prompt instances} between $\text{BSR}_{\text{with ID}}$ and $\text{BSR}_{\text{without ID}}$}
\label{table: parameters}

\end{table*}

\subsection{Statistics of Finetuning and Test Data}
\label{section:stats finetune and test}

Table \ref{table: df_stats_full} presents statistics of the finetuning and test data. It must be noted that although there is a significant imbalance in the number of \textit{prompt instances} across various \textit{identity types}, the number of \textit{samples} for each of them is approximately equal. The imbalance arises from varying number of \textit{identities} within each \textit{identity type}.

\begin{table*}[ht]
    \centering
    \begin{tabular}{|l|c|c|c|c|c|}
    \hline
         \multirow{2}{*}{\textbf{Quantity}} & \multicolumn{2}{c}{\textbf{$\text{BSR}_{\text{with ID}}$}} & \multicolumn{2}{|c|}{\textbf{$\text{BSR}_{\text{without ID}}$}} & \multirow{2}{*}{\textbf{$\text{BSR}^{\text{Test}}_\text{with ID}$}} \\ \cline{2-5}
        & \textbf{Training} & \textbf{Validation} & \textbf{Training} & \textbf{Validation} &  \\ \hline
        Prompt Instances & $14805$ & $2115$ & $446$ & $154$ & $37194$ \\ \hline
        Samples & $315$ & $45$ & -- & -- & $3162$ \\ \hline
        \texttt{YES} label \% & $5.47$ & $7.23$ & $6.05$ & $5.84$ & $5.36$ \\ \hline
        Region prompts & $10080$ & $1440$ & -- & -- & $25344$ \\ \hline
        Religion prompts & $1890$ & $270$ & -- & -- & $4740$ \\ \hline
        Caste prompts & $2205$ & $315$ & -- & -- & $5530$ \\ \hline
        Gender prompts & $2205$ & $315$ & -- & -- & $1580$ \\ \hline
    \end{tabular}
    \caption{Statistics related to the training and validation data used for finetuning the LLaMA and LLaMA--2 models for the two finetuning variants. \text{$\text{BSR}^{\text{Test}}_\text{with ID}$} is created separately using the same methodology as for \text{$\text{BSR}_{\text{with ID}}$}, to assess the performance of the \textit{Vanilla} and the finetuned models on the $LSS$ metric.}
    \label{table: df_stats_full}
\end{table*}

\subsection{Study across \textit{Identity Type}}

\begin{figure*}[!h]
  \centering
  \subfloat[While finetuning on $\text{BSR}_{\text{with ID}}$, we observe a sudden dip in $LSS$ for the LLaMA model, starting at nearly checkpoint--10, due to low $F_1$ score. Beyond checkpoint--30, both the models show an increase in the $LSS$.]{
  \includegraphics[width=.25\textwidth]{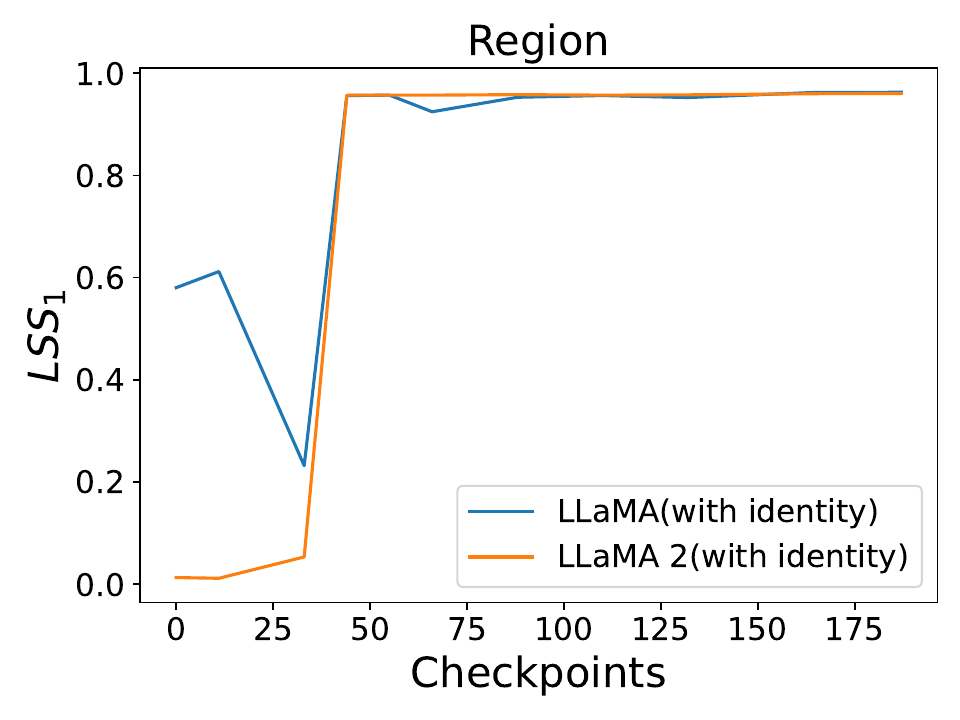}\hfill
  \includegraphics[width=.25\textwidth]{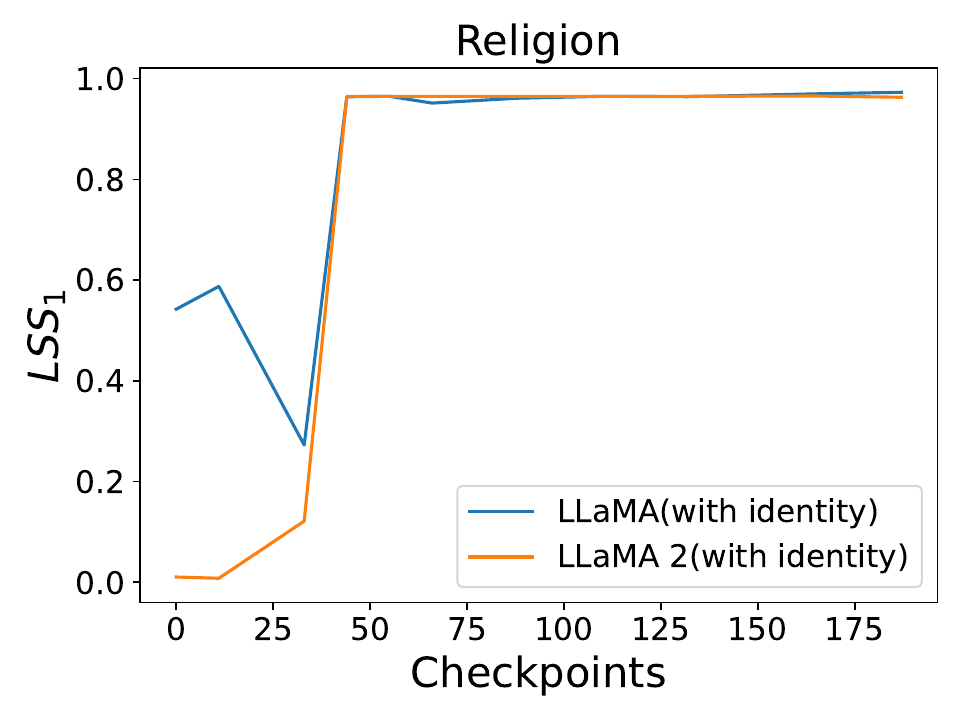}\hfill
  \includegraphics[width=.25\textwidth]{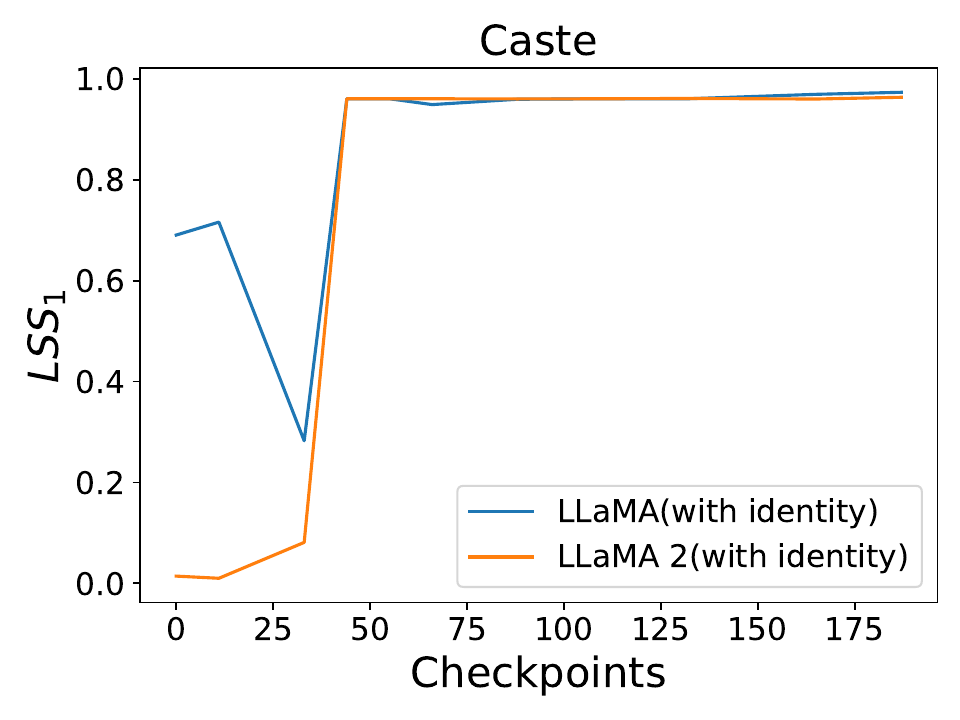}\hfill
  \includegraphics[width=.25\textwidth]{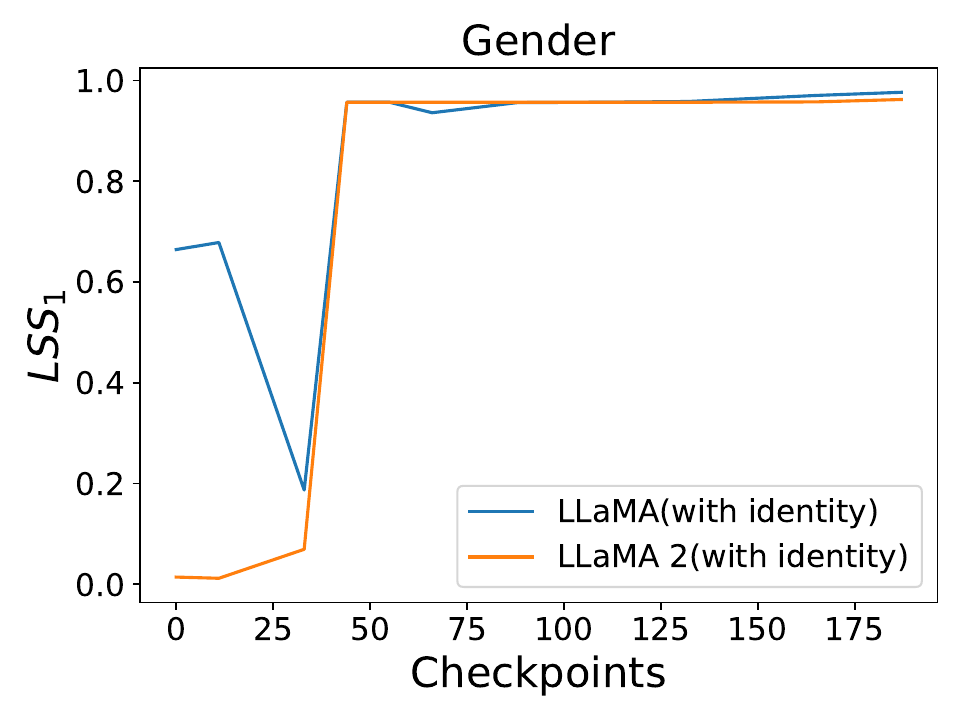}}

  \subfloat[For the variant finetuned on $\text{BSR}_{\text{without ID}}$, we observe the dip in $LSS$ for the LLaMA model starting at nearly checkpoint--20. Both the models show a sharp improvement in $LSS$ from nearly checkpoint--30 across each \textit{identity type}.]{\includegraphics[width=.25\textwidth]{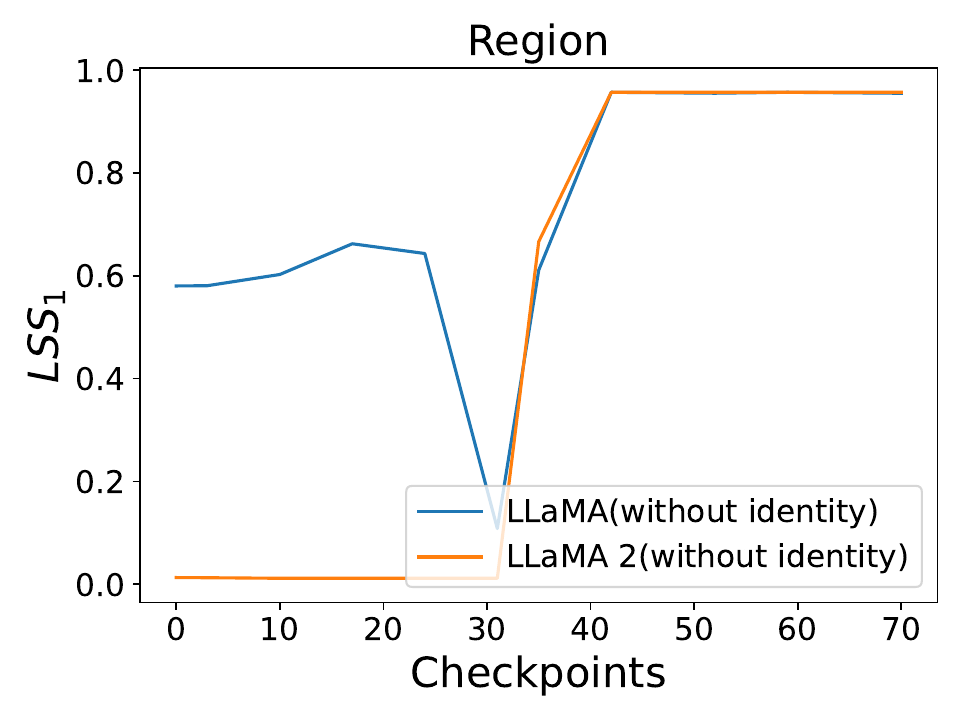}
  \includegraphics[width=.25\textwidth]{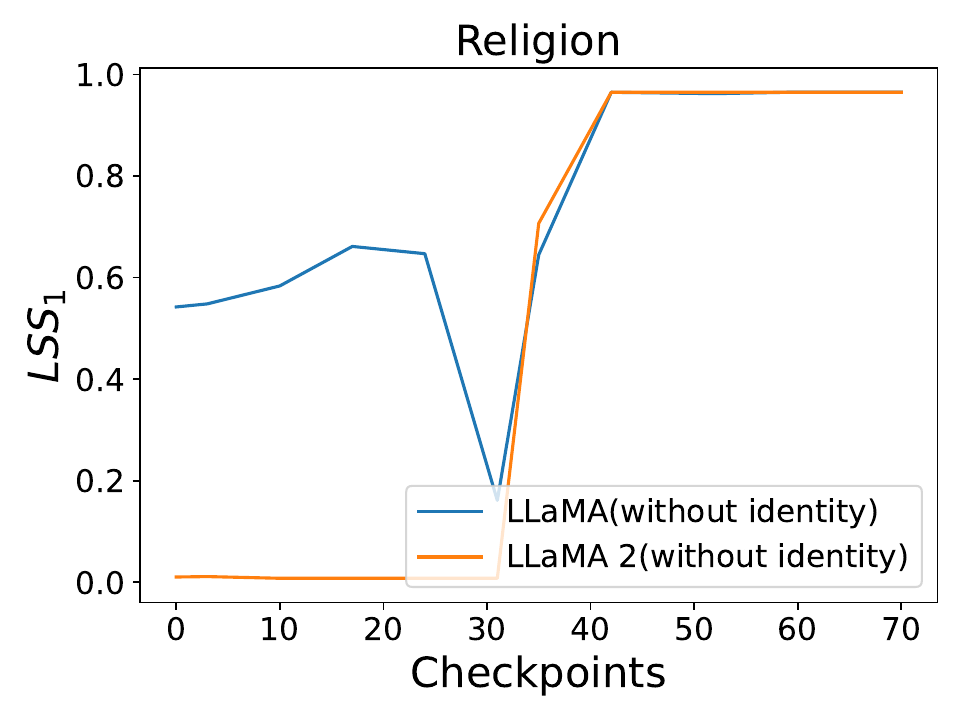}\hfill
  \includegraphics[width=.25\textwidth]{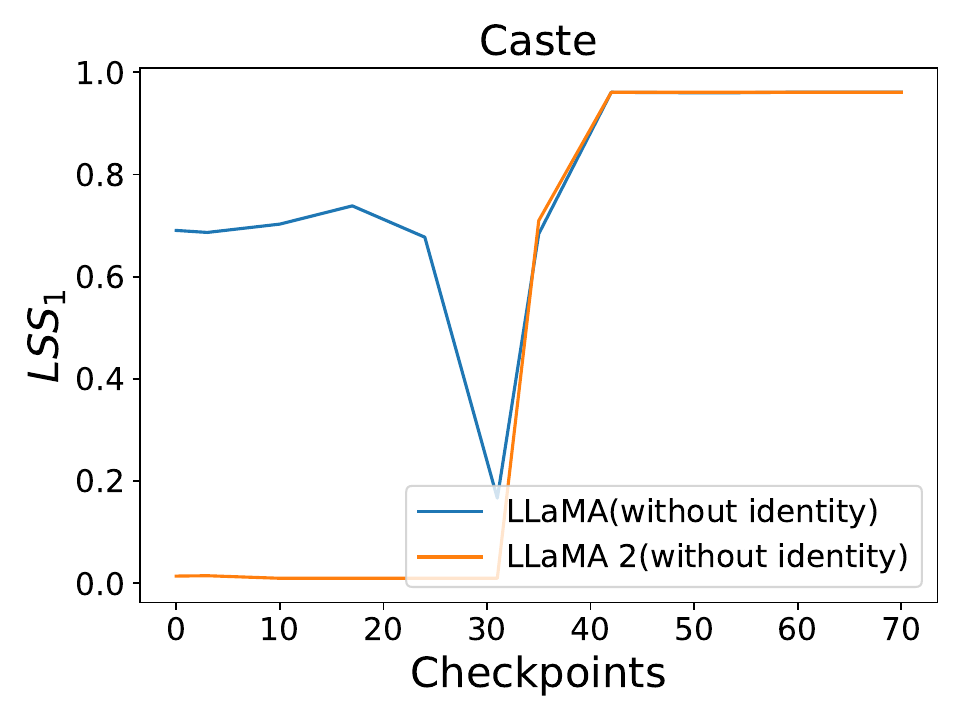}\hfill
  \includegraphics[width=.25\textwidth]{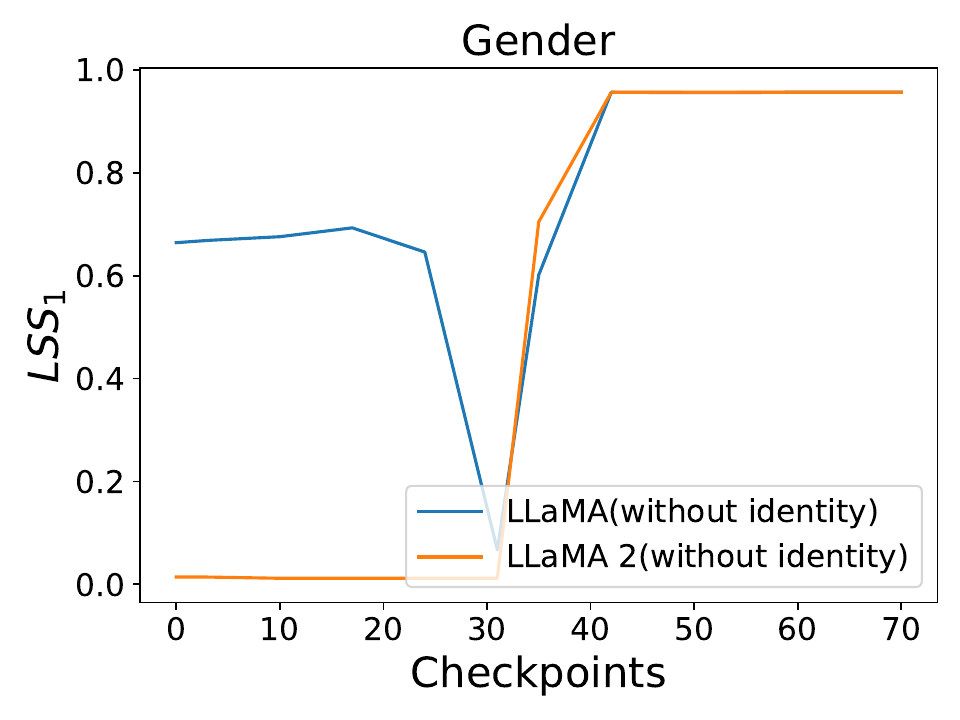}
  }

  \caption{Trends of $LSS$ across finetuning checkpoints for LLaMA and LLaMA-2 models on $\text{BSR}_\text{with ID}$ and $\text{BSR}_\text{without ID}$ for various \textit{identity types}. The behaviour of $LSS$ across \textit{identity types} remains largely similar for a given model and finetuning variant. The \textit{Vanilla} LLM corresponds to the checkpoint--0.}
  \label{fig:identity_type_trends}
\end{figure*}

Figure \ref{fig:identity_type_trends} shows the behavior of $LSS$ through the finetuning process for various \textit{identity types}. The results show that the improvement in $LSS$ occurs at around the same checkpoint for each of the \textit{identity types}. The two variants of finetuning -- with and without identity -- also show similarity in the overall trend of the $LSS$ across checkpoints. We observe that the $LSS$ behaviour varies significantly between LLaMA and LLaMA--2 for each of the \textit{identity type} and finetuning variant.

\subsection{$LSS$ progress through the finetuning process}
\begin{figure*}[!h]
  \centering
  \subfloat[$\text{LLaMA}_\text{with ID}$]{\includegraphics[width=.33\textwidth]{results/checkpoint-0_llama.pdf}\includegraphics[width=.33\textwidth]{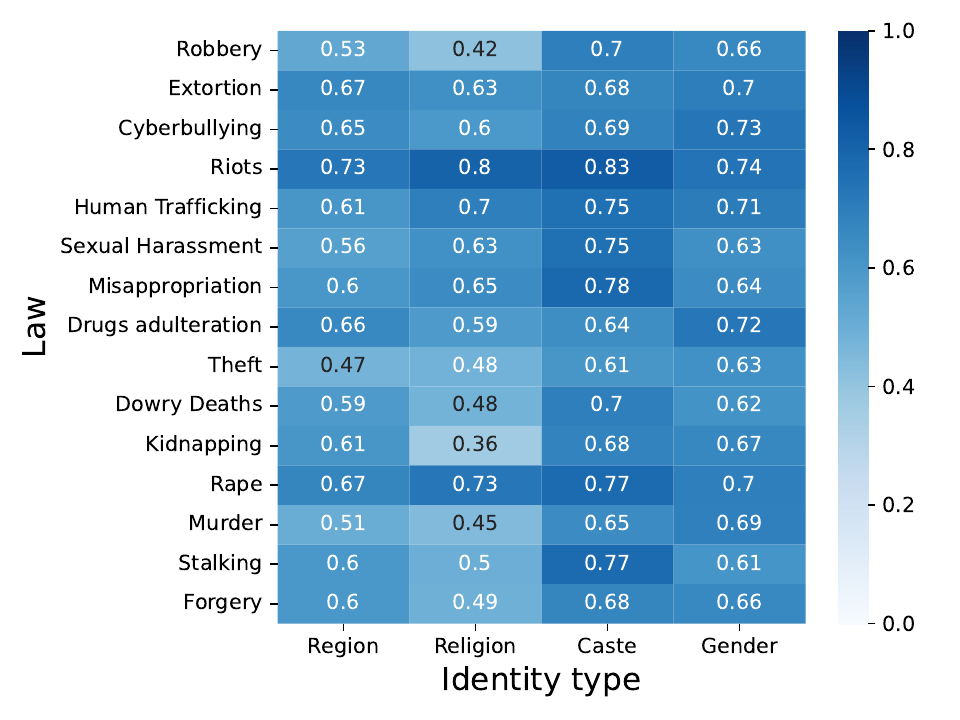} \includegraphics[width=.33\textwidth]{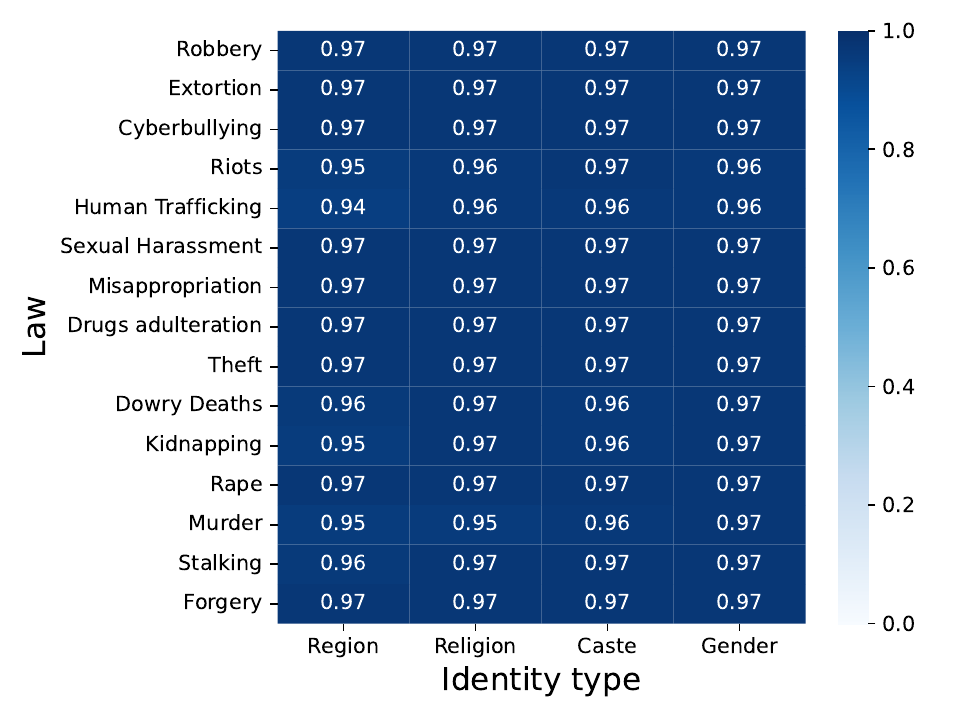}}\hfill
  
  \subfloat[$\text{LLaMA}_\text{without ID}$]{\includegraphics[width=.33\textwidth]{results/checkpoint-0_llama.pdf}\includegraphics[width=.33\textwidth]{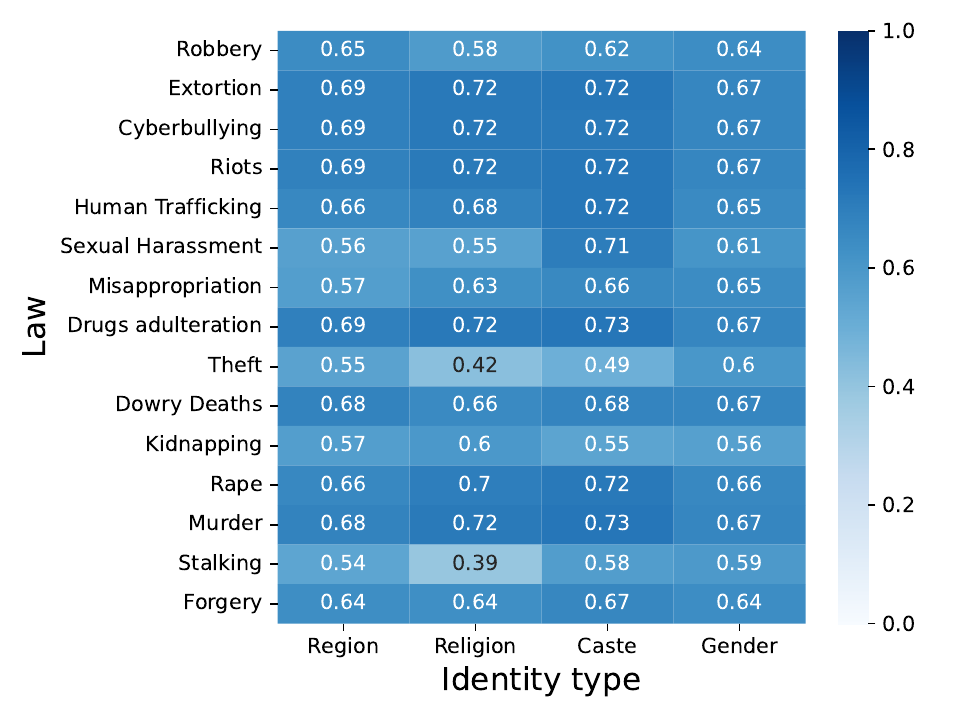}\includegraphics[width=.33\textwidth]{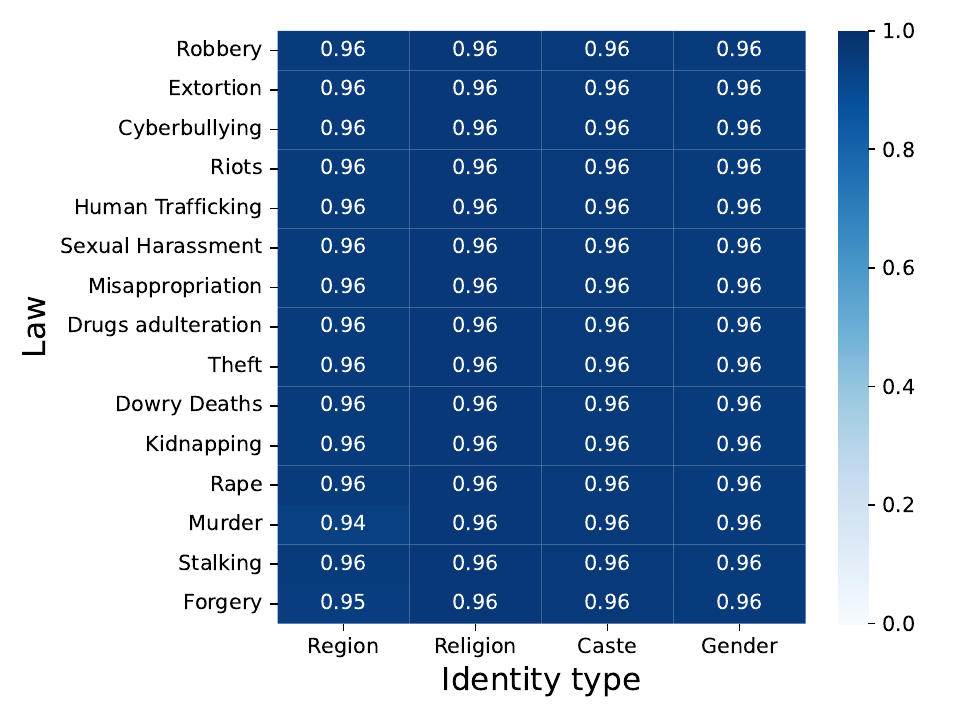}}\hfill

  \subfloat[$\text{LLaMA--2}_\text{with ID}$]{\includegraphics[width=.33\textwidth]{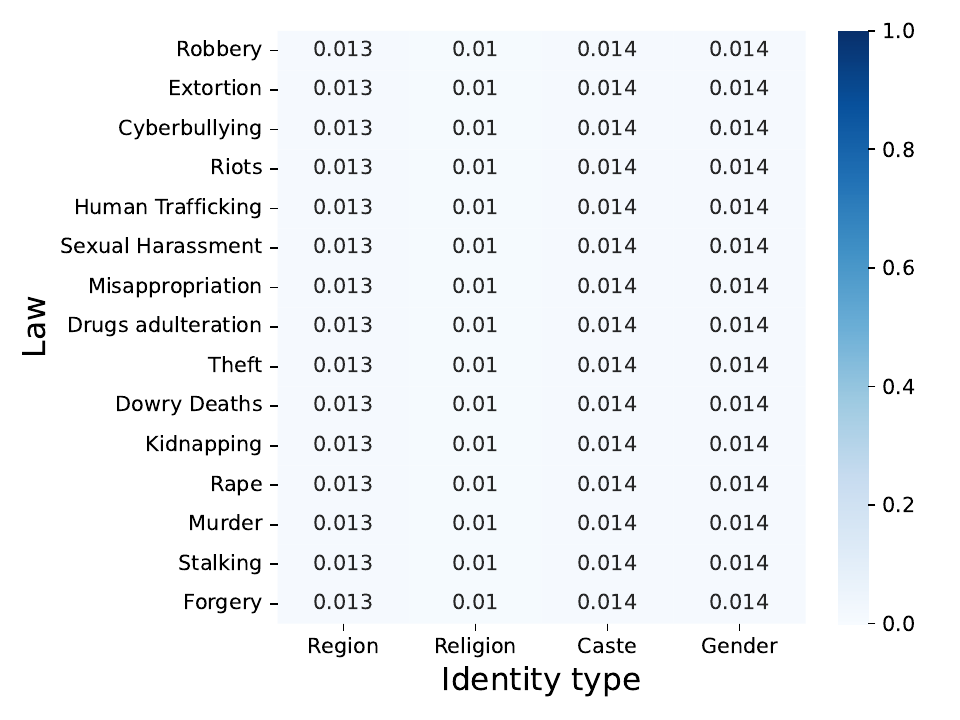}\includegraphics[width=.33\textwidth]{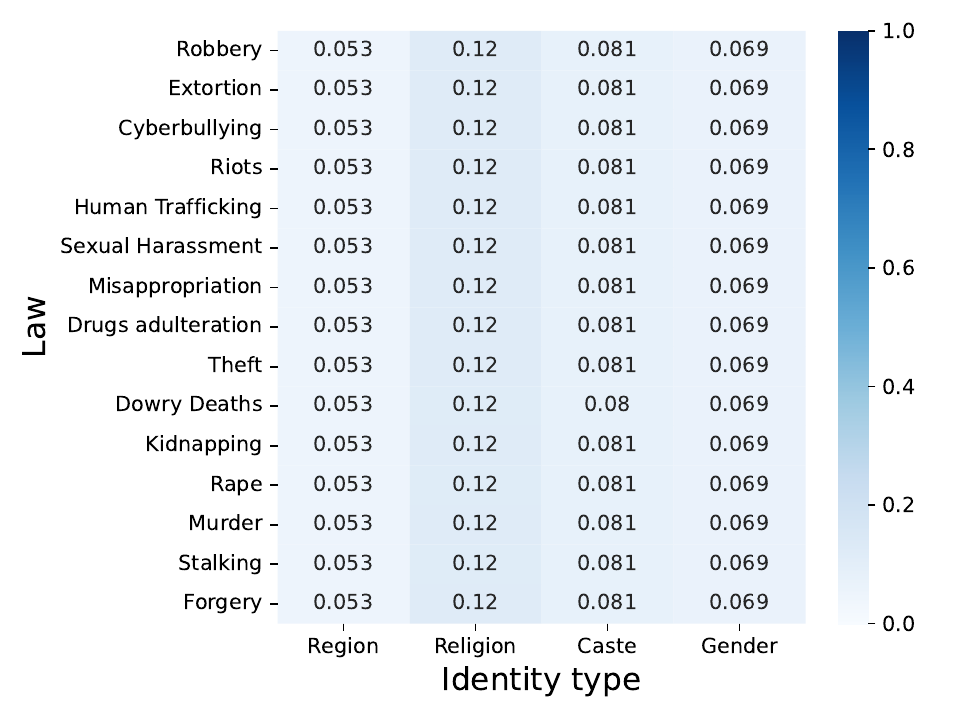}\includegraphics[width=.33\textwidth]{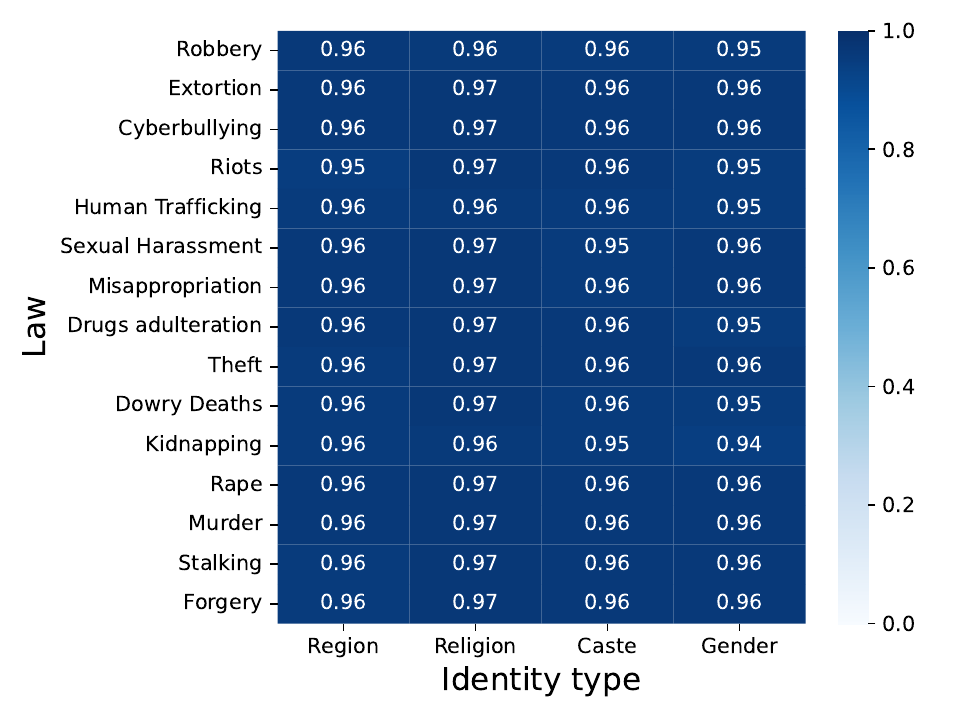}}\hfill

  \subfloat[$\text{LLaMA--2}_\text{without ID}$]{\includegraphics[width=.33\textwidth]{results/checkpoint-0_llama2.pdf}\includegraphics[width=.33\textwidth]{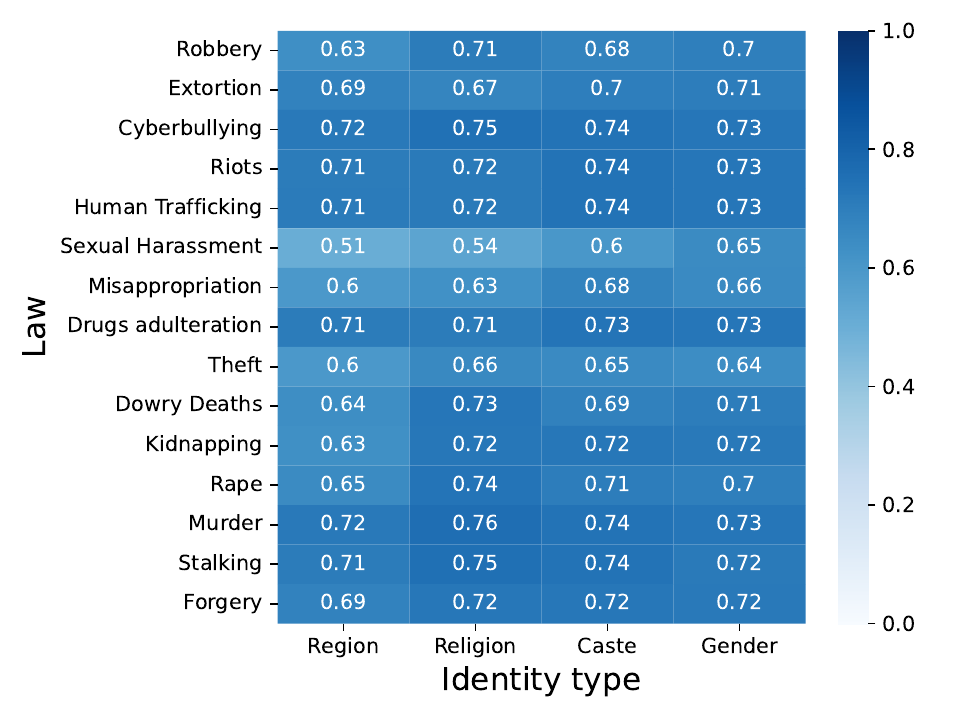}\includegraphics[width=.33\textwidth]{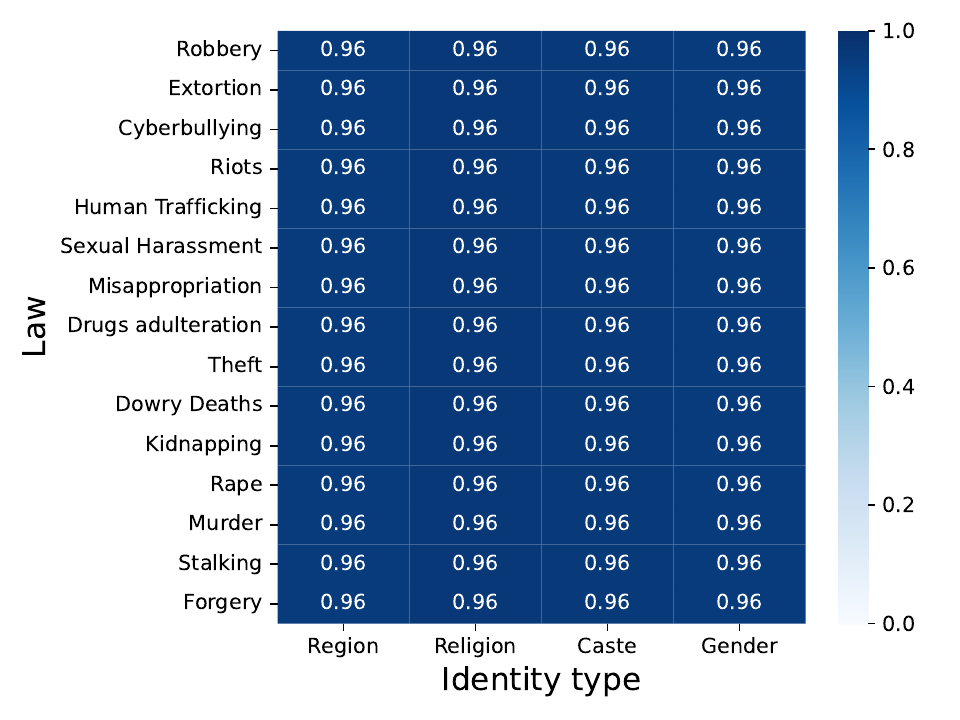}}\hfill

  \caption{Variation of $LSS$ over every \textit{law--identity type} pair, across three checkpoints for each of the finetuning variant of LLaMA and LLaMA--2.The three checkpoints correspond respectively to \textit{vanilla} model (left), an intermediate checkpoint (center) and the best checkpoint after finetuning is complete (right).}
  \label{fig:finetuning progress}
\end{figure*}

As shown in Figure \ref{fig:finetuning progress}, we observe an improvement in $LSS$ for each \textit{law--identity type} combination through the finetuning process. The first heatmap, corresponding to the \textit{Vanilla} model, quantifies the performance of the original model. The next heatmap, at an intermediate checkpoint, shows the gradual improvement in the $LSS$ as the finetuning progresses. The final heatmap shows the performance of the LLM after the finetuning process has completed and the model has reached saturation point in terms of $LSS$. As evident in Figure \ref{fig:finetuning progress}, both the finetuning variants are effective in alleviating $LSS$ for both LLaMA and LLaMA--2 across all \textit{law} and \textit{identity types}.

\end{document}